\documentclass{article}

\usepackage{times}
\usepackage{graphicx} 
\usepackage{subfigure} 

\usepackage{natbib}

\usepackage{algorithm}
\usepackage{algorithmic}

\usepackage{hyperref}

\usepackage{amsmath}
\usepackage{amssymb}
\usepackage{amsthm}
\usepackage{ifthen}
\usepackage[utf8]{inputenc}
\usepackage{url}

\usepackage{amsthm}
\usepackage{multicol}

    \theoremstyle{plain}
    \newtheorem{theorem}{Theorem}[section]
    \newtheorem{lemma}{Lemma}[section]
    \newtheorem{definition}{Definition}[section]
    \newtheorem{proposition}{Proposition}[section]
    \newtheorem{assumption}{Assumption}
    \newtheorem{corollary}{Corollary}[section]

    \newtheoremstyle{TheoremNum}
        {\topsep}{\topsep}              
        {\itshape}                      
        {}                              
        {\bfseries}                     
        {.}                             
        { }                             
        {\thmname{#1}\thmnote{ \bfseries #3}}
    \theoremstyle{TheoremNum}
    \newtheorem{thmn}{Theorem}
    
    \newtheorem{lem}{Lemma}
    
\DeclareMathOperator*{\argmin}{argmin}

\usepackage[accepted]{icml2018}

\newcommand\theTitle{On the Relationship between
Data Efficiency and Error \\
for Uncertainty Sampling}

\icmltitlerunning{On the Relationship between Data Efficiency and Error for Uncertainty Sampling}

\newcommand\sD{\ensuremath{\mathcal{D}}}

\newcommand\sI{\ensuremath{\mathcal{I}}}

\newcommand\sN{\ensuremath{\mathcal{N}}}

\newcommand\sX{\ensuremath{\mathcal{X}}}

\newcommand\R{\ensuremath{\mathbb{R}}} 
\newcommand\eqdef{\ensuremath{\stackrel{\rm def}{=}}} 

\ifthenelse{\isundefined{\definition}}{\newtheorem{definition}{Definition}}{}
\ifthenelse{\isundefined{\assumption}}{\newtheorem{assumption}{Assumption}}{}
\ifthenelse{\isundefined{\hypothesis}}{\newtheorem{hypothesis}{Hypothesis}}{}
\ifthenelse{\isundefined{\proposition}}{\newtheorem{proposition}{Proposition}}{}
\ifthenelse{\isundefined{\theorem}}{\newtheorem{theorem}{Theorem}}{}
\ifthenelse{\isundefined{\lemma}}{\newtheorem{lemma}{Lemma}}{}
\ifthenelse{\isundefined{\corollary}}{\newtheorem{corollary}{Corollary}}{}
\ifthenelse{\isundefined{\alg}}{\newtheorem{alg}{Algorithm}}{}
\ifthenelse{\isundefined{\example}}{\newtheorem{example}{Example}}{}
\newcommand{\E}{\ensuremath{\mathbb{E}}} 

\newcommand\Xu{\sX_{\text{U}}}
\newcommand\Err{\text{Err}}
\newcommand\sgn{\text{sgn}}

\newcommand\nseed{n_\text{seed}}
\newcommand\npool{n_\text{pool}}

\newcommand\npassive{n_\text{passive}}
\newcommand\nactive{n_\text{active}}

\newcommand\wpassive{\hat w_\text{passive}}
\newcommand\wactive{\hat w_\text{active}}

\newcommand\Ipassive{\sI_\text{passive}}
\newcommand\Iactive{\sI_\text{active}}

\newcommand\DE{\text{DE}}

\begin{document} 

\twocolumn[
\icmltitle{\theTitle}

\begin{icmlauthorlist}
\icmlauthor{Stephen Mussmann}{stanford}
\icmlauthor{Percy Liang}{stanford}
\end{icmlauthorlist}

\icmlaffiliation{stanford}{Stanford University, Stanford, CA}

\icmlcorrespondingauthor{Stephen Mussmann}{mussmann@stanford.edu}

\vskip 0.3in
]

\printAffiliationsAndNotice{}

\begin{abstract} 
While active learning offers potential cost savings, the actual data efficiency---the reduction in amount of labeled data needed to obtain the same error rate---observed in practice is mixed. This paper poses a basic question: when is active learning actually helpful? We provide an answer for logistic regression with the popular active learning algorithm, uncertainty sampling.
Empirically, on 21 datasets from OpenML, we find a strong inverse correlation between data efficiency and the error rate of the final classifier. Theoretically, we show that for a variant of uncertainty sampling, the asymptotic data efficiency is within a constant factor of the inverse error rate of the limiting classifier.

\end{abstract} 

\section{Introduction}

Active learning offers potential label cost savings by adaptively choosing the data points to label. Over the past two decades, a large number of active learning algorithms have been proposed
\citep{seung1992query,lewis1994sequential, freund1997selective,
tong2001support, roy2001toward, brinker2003incorporating, hoi2009semisupervised}.
Much of the community's focus is on comparing the merits of different
active learning algorithms \citep{schein2007active, yang2016benchmark}.

This paper is motivated by the observation that even for a \emph{fixed} active
learning algorithm, its effectiveness varies widely across datasets.
\citet{tong2001support} show a dataset where uncertainty sampling achieves 5x data
efficiency, meaning that active learning achieves the same error rate as
random sampling with one-fifth of the labeled data.
For this same algorithm, different datasets yield a mixed bag of results:
worse performance than random sampling \citep{yang2016benchmark}, no gains \citep{schein2007active}, gains of 2x \citep{tong2001support}, and gains of 3x \citep{brinker2003incorporating}.

In what cases and to what extent is active learning superior to naive random sampling? This is an important question to address for active learning to be effectively used in practice.
In this
paper, we provide both empirical and theoretical answers
for the case of logistic regression and uncertainty sampling, 
``the simplest and most commonly used'' active learning algorithm in practice \citep{settles2010active}
and the best algorithm given in the benchmark experiments of \citet{yang2016benchmark}.

Empirically, in Section~\ref{sec:experiments}, we study 21 binary classification datasets from OpenML. 
We found that the data efficiency for uncertainty sampling and inverse error achieved by training on the full dataset are correlated with a Pearson correlation of $0.79$ and a Spearman rank correlation of $0.67$.

Theoretically, in Section~\ref{sec:theory}, we analyze a two-stage variant of uncertainty sampling,
which first learns a rough classifier via random sampling and then samples near
the decision boundary of that classifier.
We show that the asymptotic data efficiency of this algorithm compared to random sampling
is within a small constant factor of the inverse limiting error rate.
The argument follows by comparing the Fisher information of the passive and active estimators,
formalizing the intuition that in low error regimes, random sampling wastes
many samples that the model is already confident about.
Note that this result is different in kind than the $\log(1/\epsilon)$ versus
$1/\epsilon$ rates often studied in statistical active learning theory \cite{balcan2009agnostic,hanneke2014statistical},
which focuses on convergence rates as opposed to the dependence on error. Together, our empirical and theoretical results provide a strong link between the data efficiency
and the limiting error rate.

\section{Setup}
\label{sec:setup}

Consider a binary classification problem where the goal is to learn a predictor $f$
from input $x \in \R^d$ to output $y \in \{-1,+1\}$ that has low expected error (0-1 loss),
$\Err(f) = \Pr[f(x) \neq y]$ with respect to an underlying data distribution.
In pool-based active learning, we start with a set of unlabeled input points $\Xu$.
An active learning algorithm queries a point $x \in \Xu$, receives its label
$y$, and updates the model based on $(x,y)$.
A passive learning algorithm (random sampling) simply samples points from $\Xu$ uniformly randomly without replacement, queries their labels, and trains a model on this data.

\subsection{Logistic Regression}

\label{sec:logistic_regression}
In this work, we focus on logistic regression,
where 
${p_w(y \mid x) = \sigma(yx \cdot w)}$, $w$ is a weight vector,
and ${\sigma(z) = \frac{1}{1 + \exp(-z)}}$ is the logistic function.
A weight vector $w$ characterizes a predictor ${f_{w}(x) = \sgn(x \cdot w)}$.
Given a set of labeled data points $D$ (gathered either passively or actively),
the maximum likelihood estimate is
${\hat{w} = \argmin_{w} \sum_{(x,y) \in D} - \log p_w(y \mid x)}$.
Define the limiting parameters as the analogous quantity on the population:
${w^* = \argmin_{w} \E[-\log p_w(y \mid x)]}$.
A central quantity in this work is the \emph{limiting error},
denoted ${\Err = \Err(f_{w^*})}$.
Note that we are interested in 0-1 loss (as captured by $\Err$),
though the estimator $w^*$ minimizes the logistic loss.

\subsection{Uncertainty Sampling}

In this work, we focus on ``the
simplest and most commonly used query framework'' \citep{settles2010active},
uncertainty sampling \citep{lewis1994sequential}.
This is closely related to margin-based active learning in the theoretical literature
\citep{balcan2007margin}.

Uncertainty sampling first samples $\nseed$ data points randomly from $\Xu$,
labels them, and uses that to train an initial model.
For each of the next $n - \nseed$ iterations,
it chooses
an data point from $\Xu$ that the current model is most uncertain about (i.e., closest to the decision boundary),
queries its label, and retrains the model using all labeled data points collected so far.
See Algorithm \ref{alg:us} for the pseudocode (note we
will change this slightly for the theoretical analysis).

\begin{figure}[!b]
  \begin{minipage}[t]{.98\linewidth}

    \begin{algorithm}[H]
      \caption{Uncertainty Sampling}
      \label{alg:us}
      \begin{algorithmic}
        \STATE {\bfseries Input:} Probabilistic model $p_w(y|x)$, unlabeled $X_{\text{U}}$, $\nseed$
        \STATE Randomly sample $\nseed$ points without replacement from $\sX_{\text{U}}$ and call them $\sX_{\text{seed}}$.
        \STATE $\sX_{\text{U}} = \sX_{\text{U}} \setminus \sX_{\text{seed}}$
        \STATE $\sD = \emptyset$
        \FOR{each $x$ in $\sX_{\text{seed}}$}
          \STATE Query $x$ to get label $y$
          \STATE $\sD = \sD \cup \{(x,y)\}$
        \ENDFOR
        \FOR{$n - \nseed$ iterations}
          \STATE $\hat{w} = \argmin_w \sum_{(x,y) \in \sD} -\log p_w(y|x)$
          \STATE Choose $x = \argmin_{x \in \sX_{\text{U}}} |P_{ \hat{w}}(y | x) - \frac{1}{2}|$
          \STATE Query $x$ to get label $y$
          \STATE $\sX_{\text{U}} = \sX_{\text{U}} \setminus \{x\}$
          \STATE $\sD = \sD \cup \{(x,y)\}$          
        \ENDFOR
        \STATE $\hat{w} = \argmin_w \sum_{(x,y) \in \sD} -\log p_w(y|x)$ and return $\hat{w}$
      \end{algorithmic}
    \end{algorithm}

  \end{minipage}

\end{figure}

\subsection{Data Efficiency}

Let $\wpassive$ and $\wactive$ be the two estimators obtained by performing
passive learning (random sampling) and active learning (uncertainty sampling),
respectively.
To compare these two estimators, we use \emph{data efficiency}
(also known as statistical relative efficiency \cite{vaart98asymptotic} or sample complexity ratio),
which is the reduction in number of labeled points that active learning
requires to achieve error $\epsilon$ compared to random sampling.

More precisely, consider the number of samples for each estimator to reach error $\epsilon$:
\begin{align}
  \nactive(\epsilon) &\eqdef \max\{ n : \mathbb{E}[\Err(\wactive)] \ge \epsilon \}, \\
  \npassive(\epsilon) &\eqdef \max\{ n : \mathbb{E}[\Err(\wpassive)] \ge \epsilon \},
\end{align}
where the expectation is with respect to the unlabeled pool, the labels, and
any randomness from the algorithm.
Then the data efficiency is defined as the ratio:
\begin{align}
  \DE(\epsilon) &\eqdef \frac{n_{passive}(\epsilon)}{n_{active}(\epsilon)}.
\end{align}

\subsection{Data Efficiency Dependence on Dataset}

\begin{figure}[t]
\centering
\includegraphics[width=0.99 \columnwidth]{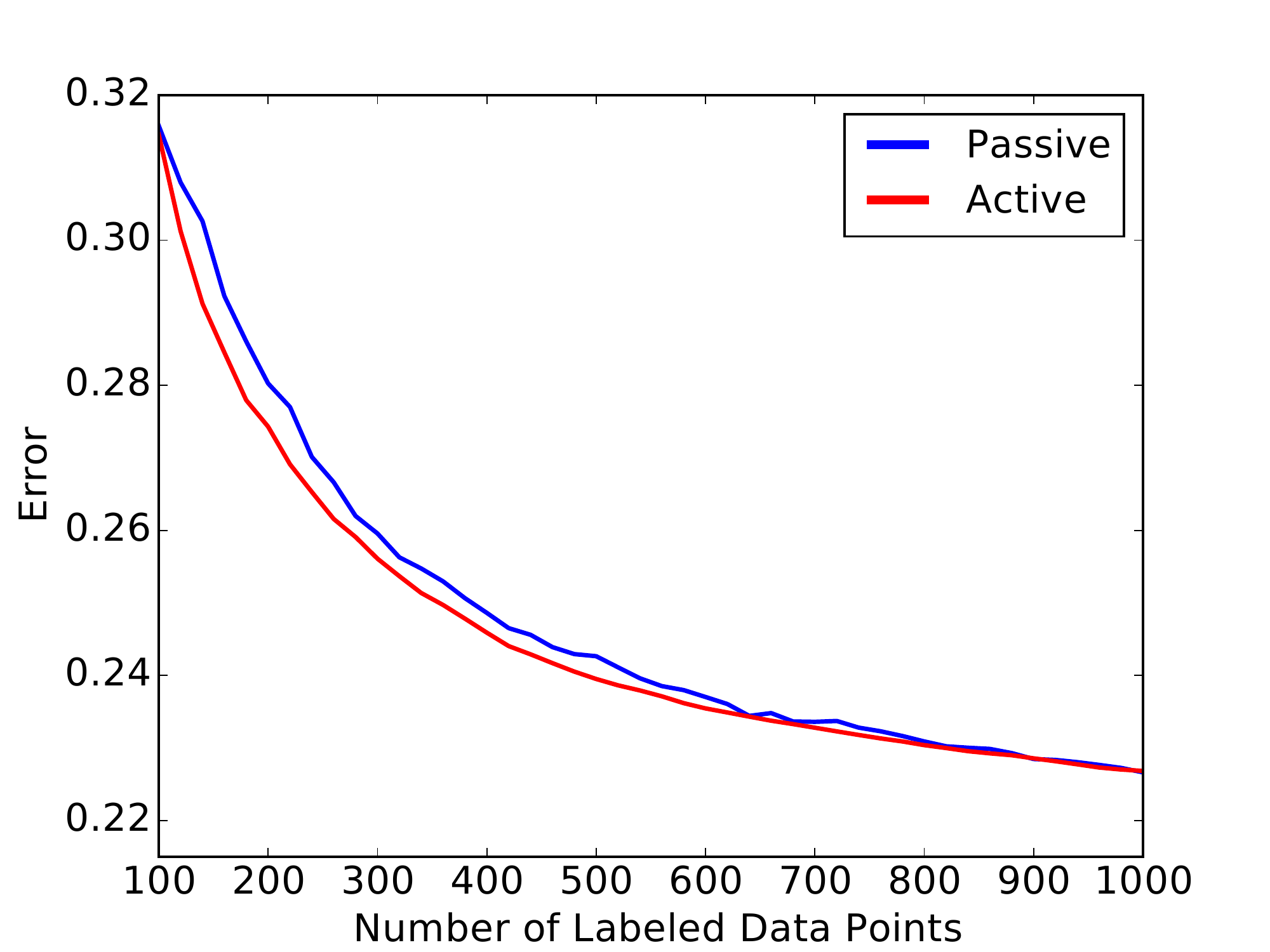}
\caption{Active learning yields meager gains when the clusters
  are closer together ($\mu=0.8$). The data efficiency is about 1x to get to 23\%
  error; both algorithms require approximately the same amount of data to achieve
  that error.}
\label{fig:bad_curve}
\end{figure}

\begin{figure}[t]
\centering
\includegraphics[width=0.99 \columnwidth]{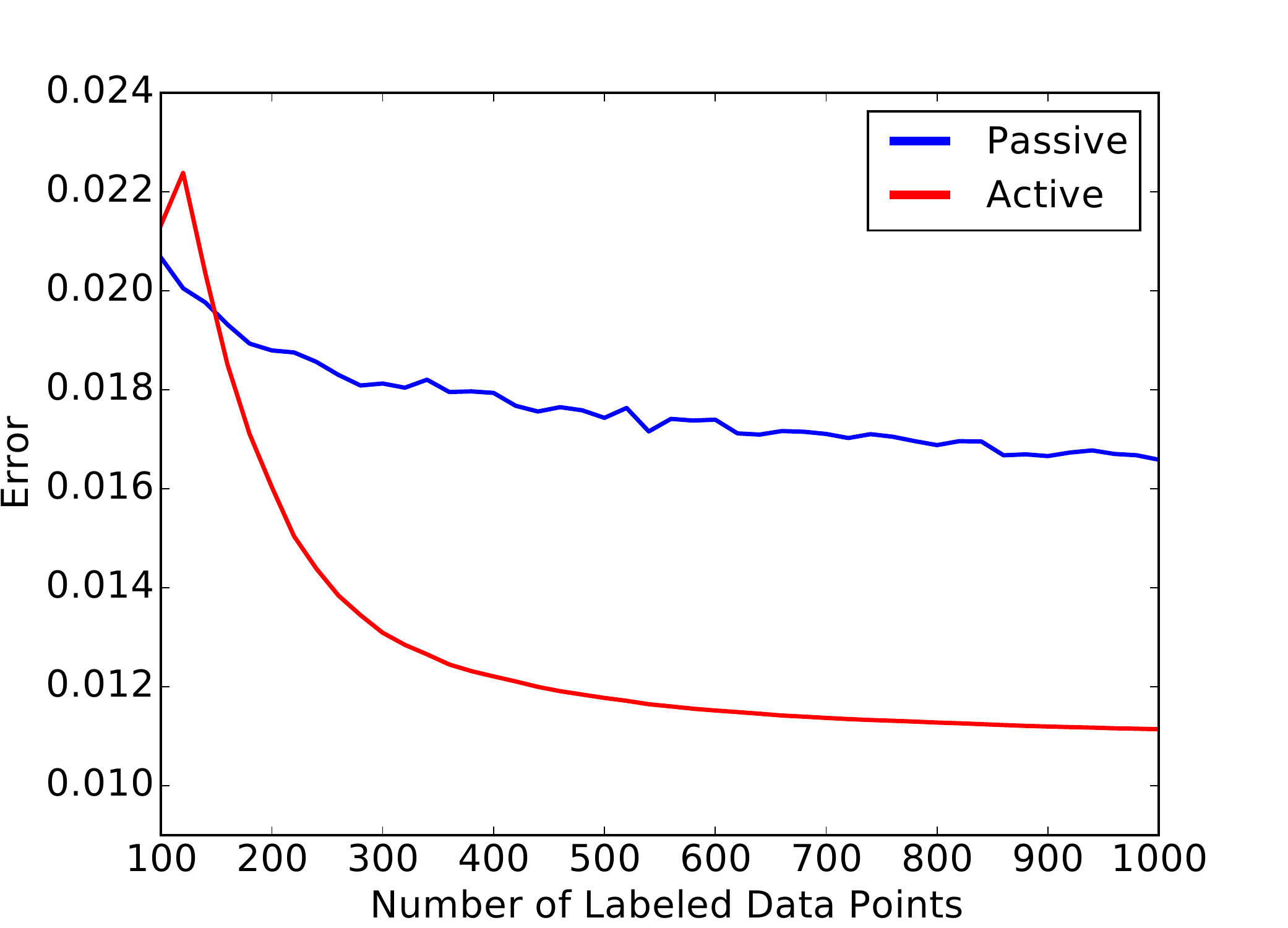}
\caption{Active learning yields spectacular gains when the
  clusters are farther apart ($\mu=2.3$). The data efficiency is about 5x to get to
  16\% error; passive learning requires about 1000 data points to achieve
  that error, while active learning only requires about 200.
  }
\label{fig:good_curve}
\end{figure}

The data efficiency depends on properties of the underlying data distribution.
In the experiments (Section \ref{sec:experiments}),
we illustrate this dependence on show a variety of real-world datasets.
As a simple illustration, we show
this phenomenon on a simple synthetic data distribution. Suppose data points
are sampled according to
\begin{align}
y \sim \text{Uniform}(\{-1,1\}), \quad x \sim \sN(y\mu e_1, I),
\end{align}
where $e_1 = [1, 0, \dots]$.
This distribution over $(x,y)$ is the standard
Gaussian Naive Bayes model with means $-\mu e_1$ and $\mu e_1$ and covariance
$I$. See Figure~\ref{fig:bad_curve} for the learning curves when $\mu=0.8$ and
Figure~\ref{fig:good_curve} for when $\mu =2.3$. We note that the data
efficiency doesn't even reach $1.1$ when $\mu=0.8$, and the curves get closer
 with more data. On the other hand, when $\mu=2.3$, the data efficiency
exceeds $5$ and increases dramatically. This illustrates the
wildly different gains of active learning, depending on the dataset.
In particular, the data efficiency is higher for the less noisy
dataset, as the thesis of this work predicts.

\section{Experiments}
\label{sec:experiments}

\subsection{Datasets}

We wish to study the data efficiency of active learning versus passive learning
across a comprehensive set of datasets which are ``typical'' of real-world settings.
Capturing a representative set of datasets is challenging,
and we wanted to be as objective and transparent about the process as possible,
so we detail the dataset selection process below.

We curated a set of datasets systematically from OpenML,
avoiding synthetic or degenerate cases.
In August 2017, we downloaded all 7968 datasets.
We removed datasets with missing features or over 1 million data points.
We wanted a large unlabeled pool (relative to the number of features)
so we kept datasets where
the number of features was less than 100 and the number of data points was at least 10,000.
In our experiments, we allow each algorithm to query the label of $n=1000$ points,
so this filtering step ensures that $d \le n/10$ and $\npool \ge 10 n$.
We remark that more than 98\% of the datasets were filtered out because they
were too small (had fewer than 10,000 points).
138 datasets remained. 

We further removed datasets that were synthetic, had unclear descriptions, or were duplicates.
We also removed non-classification datasets.
For multiclass datasets, we processed them to binary classification by predicting
majority class versus the rest.
Of the 138 datasets, 36 survived.

We ran standard logistic regression on training splits of these datasets.
In 11 cases, logistic regression was less than 1\%
better than the classifier that always predicts the
majority class. Since logistic regression was not meaningful for these
datasets, we removed them, resulting in 25 datasets.

On one of these datasets, logistic regression
achieved 0\% error with fewer than $40$ data points.
On another dataset,
the performance of random sampling became \emph{worse} as the number of labels increased. On two datasets, active learning achieved at least 1\% error
lower than the error with the \emph{full training set}, a phenomenon that
\citet{schohn2000less} calls ``less is more''; this is beyond the
scope of this work. We removed these four cases,
resulting a total of 21 datasets.

The final 21 datasets has a large amount of variability, from
healthcare, game playing, control, ecology, economics, computer vision,
security, and physics. 

\subsection{Methodology}

We used a random sampling seed of size $\nseed = 100$ and plotted the learning curves up
until a labeling budget of $n = 1000$. We calculated the data efficiency at the
lower of the errors achieved with the $n = 1000$ budget by active and passive learning.
As a proxy for the limiting error, we use the error on the test set obtained by a classifer
trained on the full training set.

\subsection{Results}

\begin{figure}
\centering
\includegraphics[width=0.99 \columnwidth]{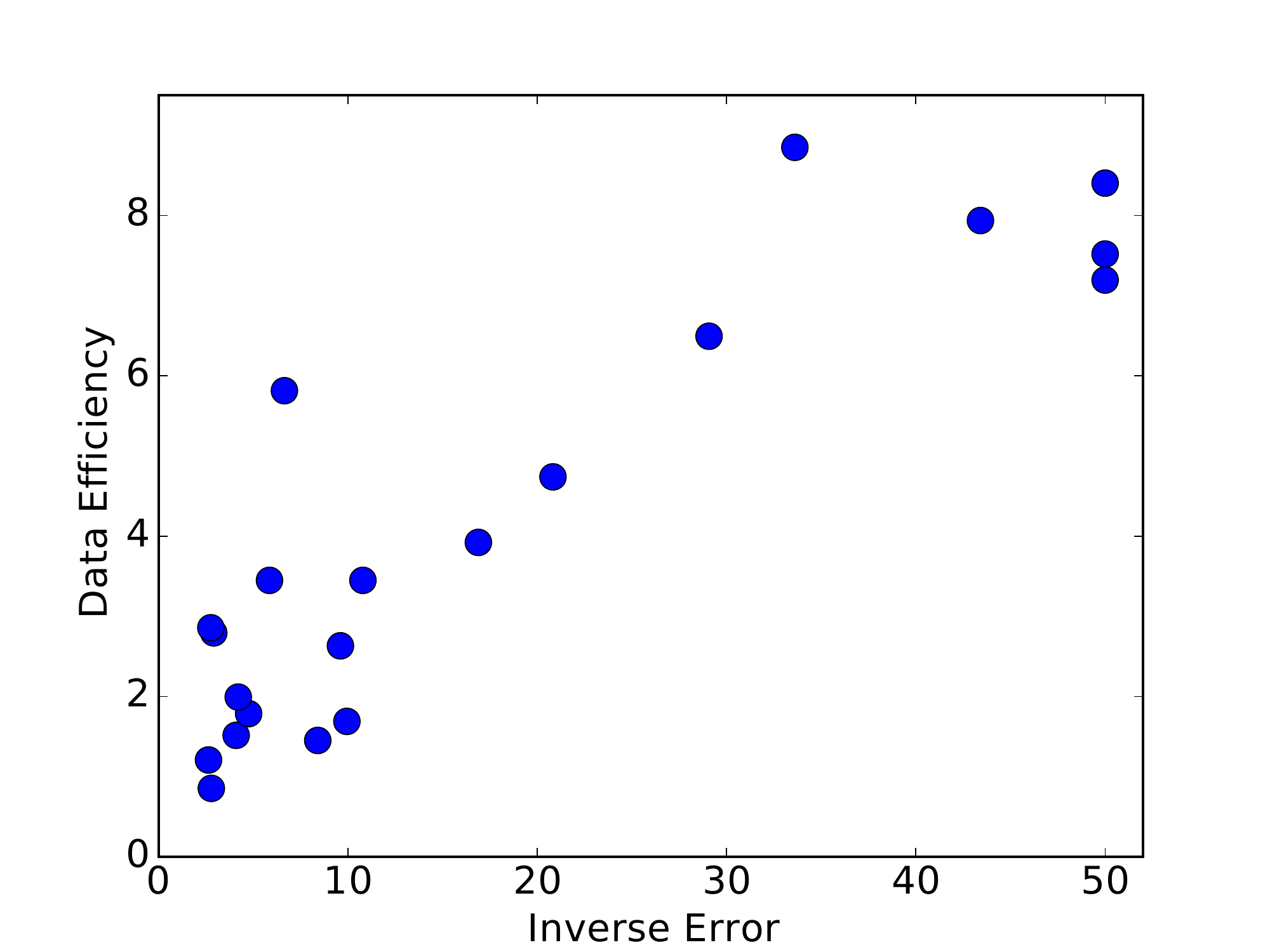}
\caption{Scatterplot of the data efficiency of uncertainty sampling versus the inverse error using all the data. Line of best fit has 0.789 $R^2$, also known as the Pearson Correlation.}
\label{fig:deie}
\end{figure}

Figure~\ref{fig:deie} plots the relationship between data efficiency and
the inverse error across all datasets.
To remove outliers, we capped the inverse error at $50$;
this truncated the inverse error of three datasets which had inverse error of
$190$, $3200$, and $27000$ which corresponds to errors less than around 0.5\%. 
The correlation ($R^2$ of the best linear fit) is $0.789$. Further, the data efficiency and the inverse error have a
Spearman rank correlation of $0.669$.

In summary, we note that data efficiency is closely tied to the inverse error.
In particular, when the error is below 10\%, the data efficiency is at least 3x
and can be much higher.

\section{Theoretical Analysis}
\label{sec:theory}

In this section, we provide theoretical insight into the
inverse relationship between data efficiency and limiting error.
For tractability,
we study the asymptotic behavior as the number of labels $n$ tends to infinity.

Let $p(x,y)$ be the underlying data distribution over $\mathbb{R}^d \times \{-1,1\}$.
For uncertainty sampling, there are three data quantities:
$\nseed$, the number of seed data points;
$n$, the amount of labeled data (the budget); and
$\npool$, the number of unlabeled points in the pool.
We will assume that $\nseed$ and
$\npool$ are functions of $n$, and we let $n$ go to infinity.
In particular, we wish to bound
the value of $\lim_{\epsilon \rightarrow \Err} \DE(\epsilon)$ where $\Err$ is
the limiting error defined in Section \ref{sec:logistic_regression}.
Bounding $\DE(\epsilon)$ for small $\epsilon$ is closely related to the statistical
asymptotic relative efficiency \citep{vaart98asymptotic}. We use data efficiency as it applies for finite $n$.

The asymptotic data efficiency only makes sense if the random
sampling and uncertainty sampling both converge to the same error. Otherwise,
the asymptotic data efficiency would either be $0$ or $\infty$. While bias in active
learning is an important topic of study \cite{liu2015shift}, it is beyond the
scope of this work. We will make an assumption that ensures this is
satisfied if the model is well-specified in some small slab around the decision
boundary. 

\subsection{Two-stage Variant of Uncertainty Sampling}

Because of the complicated coupling between uncertainty sampling
draws and updates, we analyze a two-stage variant: we gather an initial seed
set using random sampling from the unlabeled dataset,
and then gather the points closest to the
decision boundary learned from the seed data.
This two-stage approach is similar to other active learning
work \citep{chaudhuri2015convergence}. 

Thus, we only update the parameters twice: after the seed round we train on the seed data, and after we have collected all the data, we train on the data that was collected after the seed data. We do not update the parameters between draws closest to the decision boundary.

Also, instead of always choosing the point closest to decision boundary without replacement during the uncertainty sampling phase, with $\alpha>0$ probability we randomly sample from the unlabeled pool and with $1-\alpha$ probability we choose the point closest to the decision boundary. The random sampling proportion $\alpha$ ensures that the empirical data covariance is non-singular for uncertainty sampling.

\subsection{Sketch of Main Result}

Under assumptions that will be described later, our main result is that there exists some
$\epsilon_0$ such that for any $\Err < \epsilon < \epsilon_0$,
\begin{align}
DE(\epsilon) > \frac{s}{4 \Err},
\end{align}
where $s$ is a constant bounding a ratio of conditional covariances in the
directions orthogonal to $w^*$. In particular, if the pdf factorizes into two
marginal distributions (decomposition of $x$ into two independent components),
one along the direction of $w^*$ and one in the directions orthogonal to $w^*$,
then the conditional covariances orthogonal to $w^*$ are equal, and $s=1$. 
If the distribution is additionally symmetric across the decision
boundary, we obtain
\begin{align}
\frac{1}{4 \Err} < DE(\epsilon) < \frac{1}{2 \Err}.
\end{align}

We now give a rough proof sketch.
The core idea is to compare the Fisher information of active and
passive learning, similar to other work in the literature
\citep{sourati2017asymptotic}. It is known that the Fisher information matrix
for logistic regression is
\begin{align}
\sI= \mathbb{E}[\sigma (1-\sigma) xx^\top],
\end{align}
where $\sigma = \sigma(yx \cdot w^*)$.
Note that $\sigma$ only depends on the part of $x$ parallel to $w^*$. If the data decomposes into two independent components as mentioned above, then
\begin{align}
\Ipassive = \mathbb{E}[\sigma (1-\sigma)] \mathbb{E}[xx^\top]
\end{align}
if we ignore the dimension of the Fisher information along $w^*$ which doesn't end up mattering (it only changes the magnitude of $w^*$ which is independent of the 0-1 loss). Additionally, since uncertainty sampling samples at the decision boundary where $w \cdot x^*=0$,
we have $\sigma = \frac{1}{2}$ and thus active learning achieves:
\begin{align}
\Iactive = \frac{1}{4} \mathbb{E}[xx^\top].
\end{align}
The Fisher information determines the asymptotic rate of convergence of the parameters:
\begin{align}
  \sqrt{n}(w_n - w^*) \stackrel{d}{\rightarrow} \sN(0,\sI^{-1}).
\end{align}
Intuitively, this convergence rate is monotonic with $\sI^{-1}/n$ which means the ratio (abuse of notation, but true for any linear function of the inverse Fisher information) of the inverse Fisher information matrices gives the asymptotic relative rate,

\begin{align}
\DE(\epsilon) \approx \frac{\Ipassive^{-1}}{\Iactive^{-1}} \approx \frac{ 1/4 }{ \mathbb{E}[\sigma (1-\sigma)]}.
\end{align}

If the optimal logistic model is \emph{calibrated}, meaning the model's predicted probabilities are on average correct, then
\begin{align}
\frac{\Err}{2} \leq \mathbb{E}[\sigma (1-\sigma)] \leq \text{\Err}.
\end{align}

Putting these together, we get:
\begin{align}
\frac{1}{4 \Err} \lessapprox DE(\epsilon) \lessapprox \frac{1}{2 \Err}.
\end{align}

Having given the rough intuition, we now go through the arguments more formally.

\subsection{Notation}
Let $w^*$ be the limiting parameters,
Let $w_0$ be the weights after the seed round for active learning,
and $w_n$ be the weights at the end of learning with $n$ labels.

We include a bias term for logistic regression by inserting a coordinate at the beginning of $x$ that is always $1$. Thus, $x_0=1$ and $w^*_0$ is the bias term of the optimal parameters.
As a simplification of notation, the pdf $p(x)$ is only a function of the non-bias coordinates (otherwise, such a pdf wouldn't exist).

Since logistic regression is invariant to translations (we can appropriately change the bias) and rotations (we can rotate the non-bias weights), without loss of generality, we will assume that $w^* = \|w^*\| e_1$, that the bias term is $0$, and that the data is mean $0$ for all directions orthogonal to $w^*$, ($\mathbb{E}[x_{2:}]=0$).

\subsection{Assumptions}

We have four types of assumptions: assumptions on the values of $\nseed$ and $\npool$, assumptions on the distribution of $x$, assumptions on the distribution of $y$, and non-degeneracy assumptions. As an example, all these assumptions are satisfied if $\nseed = \sqrt{n}$, $\npool = n \sqrt{n}$, $x$ is a mixture of truncated, mollified Gaussians, and $y$ is well-specified for non-zero weights.

\subsubsection{Assumptions relating $\nseed,n,\npool$}

Recall that $\nseed$ is the number of labels for the seed round, $n$ is the labeling budget, and $\npool$ is the number of unlabeled data points.

\begin{assumption}[Data Pool Size]
\label{assum:pool}
$\npool = \omega(n)$.
\end{assumption}

\begin{assumption}[Seed Size]
\label{assum:seed}
$\nseed = \Omega(n^\rho)$ for some $\rho>0$ and $\nseed = o(n)$.
\end{assumption}

We need the size $\npool$ of the unlabeled pool has to be large enough so that
uncertainty sampling can select points close to the decision boundary. We
require that the seed for uncertainty sampling is large enough to make the
decision boundary after the seed round converge to the true decision boundary,
and we require that it is small enough so that it doesn't detract from the
advantages of uncertainty sampling. 

\subsubsection{Assumption on $x$ distribution}

We assume that the distribution on $x$ has a pdf (``continuous distribution''), and the following two conditions hold:

\begin{assumption}[Bounded Support]
\label{assum:bounded_support}
\begin{align}
\exists B>0: \Pr[\|x\| > B] = 0
\end{align}
\end{assumption}

\begin{assumption}[Lipschitz]
\label{assum:lipschitz}
The pdfs and conditional pdfs $p(x), p(x|w \cdot x=b), p(x|w_1 \cdot x_1=b_1, w_2 \cdot x_2=b_2)$ are all Lipschitz.
\end{assumption}

\subsubsection{Assumptions on $x,y$ distribution}

These next three assumptions (Assumptions \ref{assum:convergence}--\ref{assum:calibrate}) are implied if the logistic regression model is
well-specified ($\Pr[y|x] = \sigma(y x \cdot w^*)$),
but they are strictly weaker.
If the reader is willing to assume well-specification, this section can be skipped.

\begin{assumption}[Local Expected Loss is Zero]
\label{assum:convergence}
There exists $\lambda$ such that for $\|w - w^*\| \leq \lambda$,
\begin{align}
\mathbb{E}_{w \cdot x = 0}[\nabla_w (- \log p_{w^*}(x,y))] = 0
\end{align}
\end{assumption}

Assumption \ref{assum:convergence} is satisfied if model is well-specified in any thin slab around the decision boundary defined by $w^*$. We need this assumption to conclude that our two-stage uncertainty sampling algorithm converges to $w^*$.

\begin{assumption}[Conditions on Zero-One Loss]
\label{assum:z_conditions}
Let $Z(w) = \Pr_{x,y}[yx \cdot w < 0]$ be the zero-one loss of the classifier defined by the weights $w$. Then,
\begin{itemize}
\item $Z$ is twice-differentiable at $w^*$,
\item $Z$ has a local optimum at $w^*$, and
\item $\nabla^2 Z(w^*) \neq 0$.
\end{itemize}
\end{assumption}

In order to conclude that convergence to the optimal parameters implies convergence in error, we need Assumption \ref{assum:z_conditions}.
The strongest requirement is the local optimum part.
The twice differentiable condition is a regularity condition and the Hessian condition is generically satisfied.

\begin{assumption}[Calibration]
\label{assum:calibrate}
\begin{align}
\Pr[y| w^* \cdot x = a] = \sigma(y a)
\end{align}
\end{assumption}

\begin{figure}
\centering
\includegraphics[width=0.95 \columnwidth]{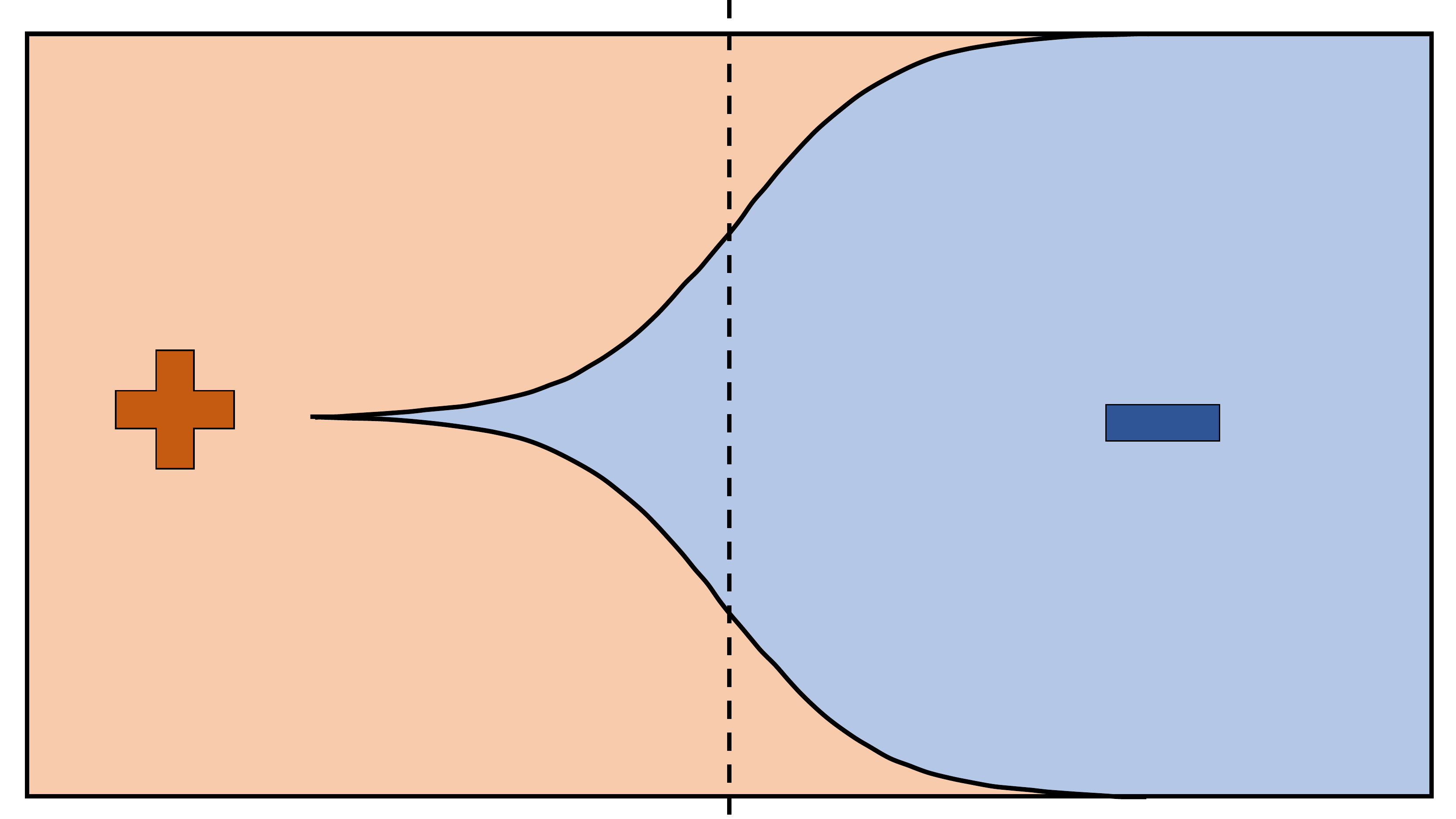}
\caption{Example of distribution with deterministic labels which is calibrated but not well-specified for logistic regression.}
\label{fig:deterministic_distribution}
\end{figure}

We say a model is \emph{calibrated} if the probability of a class, conditioned on the model predicting probability $p$, is $p$. Assumption \ref{assum:calibrate} amounts to assuming that the logistic model with the optimal parameters $w^*$ is calibrated. Note that this is significantly weaker than assuming that the model is well-specified ($\Pr[y|x] = \sigma(y x \cdot w^*)$). For example, the data distribution in Figure \ref{fig:deterministic_distribution} is calibrated but not well-specified.

These three assumptions all hold if the logistic distribution is well-specified, meaning $\Pr[y|x] = \sigma(y x \cdot w^*)$.

\subsubsection{Non-degeneracy}

Define $p_0 = \int_{w^* \cdot x = 0} p(x)$ as the marginal probability \emph{density} of selecting a point at the decision boundary. More precisely, $p_0$ is the probability density $p(x)$ integrated over the $d-1$ dimensional hyperplane manifold defined by $w^* \cdot x=0$. Equivalently, $p_0$ is the probability density of the random variable $\frac{w^*}{\|w^*\|} \cdot x$ at $0$.

\begin{assumption}[Non-degeneracies]
\label{assum:non_singular}
\label{assum:non_zero_prob}
\begin{align}
p_0 \neq 0, \quad
\|w^*\|\neq 0, \quad
\Err \neq 0, \quad
\det(\mathbb{E}[xx^\top]) \neq 0
\end{align}
\end{assumption}
Let us interpret these four conditions.
We assume that the probability \emph{density} at the decision boundary is non-zero, $p_0 \neq 0$, otherwise uncertainty sampling will not select points close to the decision boundary (note this is not an assumption about the probability \emph{mass}). We assume that $\|w^*\| \neq 0$, meaning that the classifier is not degenerate, with all points on the decision boundary. We assume $\Err \neq 0$, meaning the logistic parameters do not achieve 0\% error. Finally, we assume $\det(\mathbb{E}[xx^\top]) \neq 0$, meaning that the data covariance is non-singular, or equivalently, that the parameters are identifiable.

\subsection{Proofs}

We will first prove a condition on the convergence rate of the
error based on a quantity $\Sigma$ closely related to the Fisher Information.
However, we can't rely on the usual Fisher information analysis, which does
not connect to the zero-one loss, but rather to the asymptotic normality of the
parameters. Thus, our conditions for this key lemma are slightly stronger than
the asymptotic normality result of Fisher Information.

\subsubsection{Rates Lemma in Terms of $\Sigma$}

The logistic loss (negative log-likelihood) for a single data point under logistic regression is
\begin{align}
\ell_w(x,y) = \log(1 + \exp(-w \cdot yx)).
\end{align}
Further, the gradient and Hessian are,
\begin{align}
\nabla \ell_w(x,y) &= - yx \sigma(-w \cdot yx) \\
\nabla^2 \ell_w(x,y) &= \sigma(w \cdot yx) \sigma(-w \cdot yx) xx^\top
\end{align}
Note that $\sigma(-x) = 1-\sigma(x)$.

Following the Fisher Information asymptotic normality analysis, note that
\begin{align}
\sqrt{n} (w_n - w^*) = A_n^{-1} b_n,
\end{align}
where 
\begin{align}
A_n &= \frac{1}{n} \sum_i \nabla^2 \ell_{w'}(x_i,y_i), \\
b_n &= \frac{1}{\sqrt{n}} \sum_i \nabla \ell_{w^*}(x_i,y_i),
\end{align}
with $\|w' - w^*\| \leq \|w_n - w^*\|$. This is justified by Taylor's theorem since the logistic loss is smooth.

From these, we can define the key quantity $\Sigma$ that is equivalent to the inverse Fisher Information under stronger conditions.
\begin{definition}
\label{def:sigma}
  If $A_n \stackrel{P}{\rightarrow} A$ (non-singular and symmetric) and $\mathbb{E}[b_n b_n^T] \rightarrow B$ exists, then define
\begin{align}
\Sigma = A^{-1} B A^{-1}.
\end{align}
\end{definition}
This quantity is important because of the following lemma, which translates comparisons in the asymptotic variances to comparisons of data efficiency. Recall that without loss of generality, we let $w^* = \|w^*\|e_1$. Define $A_{-1}$ as the matrix $A$ without the first row and column.
\begin{lemma}
\label{lem:rates}
If we have two estimators with asymptotic variances $\Sigma_a$ and $\Sigma_b$, and for any $\epsilon>0$ and both estimators, $n \Pr[\|A_n - A\| \geq \epsilon] \rightarrow 0$ and $n \Pr[\|w_n - w^*\| \geq \epsilon] \rightarrow 0$, then 
\begin{align}
\Sigma_{a,-1} \succ c \Sigma_{b,-1}
\end{align}
implies that for some $\epsilon_0$ and any $\Err<\epsilon<\epsilon_0$,
\begin{align}
n_a(\epsilon) \geq c n_b(\epsilon).
\end{align}
\end{lemma}

The proof is in the appendix. This lemma only requires Assumption \ref{assum:z_conditions}, the condition on $Z$ at $w^{*}$, and is possibly of independent interest.

Note that with the bias term, our weight vector is $d+1$ dimensional, so $\Sigma$ is a square $d+1$ dimensional matrix. However, without the first row and column, $\Sigma_{-1}$ is a square $d$ dimensional matrix. The fact that the rates depend on $\Sigma_{-1}$ instead of $\Sigma$ is necessary for our results. Intuitively, the first coordinate (in direction of $w^*$) has slow convergence for uncertainty sampling since we are selecting points near the decision boundary which have small projection onto $w^*$ and thus we gain little information about the dependence of $y$ on $x_1$. However, because our analysis is in terms of the convergence of the 0-1 error rather than convergence of the parameters, the above lemma doesn't depend on the convergence rate of the first coordinate.

From this lemma, it follows that if
\begin{align}
c_1 \Sigma_{\text{active},-1} \prec \Sigma_{\text{passive},-1} \prec c_2 \Sigma_{\text{active},-1},
\end{align}
then for sufficiently small error,
\begin{align}
c_1 \leq DE(\epsilon) \leq c_2.
\end{align}

\subsubsection{Specific Calculations for Algorithms}

In proving the later results, it's useful to first establish the consistency of our algorithms. Assumption \ref{assum:convergence} is used here.

\begin{lemma}
\label{convergence}
Both two-stage uncertainty sampling and random sampling converge to $w^*$.
\end{lemma}

Next, we need our two algorithms satisfy the conditions of Lemma \ref{lem:rates}.

\begin{lemma}
\label{lem:ck_conditions}
For our active and passive learning algorithms,
for any $\epsilon>0$, $n \Pr[\|A_n - A\| \geq \epsilon] \rightarrow 0$ and $n \Pr[\|w_n - w^*\| \geq \epsilon] \rightarrow 0$.
\end{lemma}

Now, we are ready for the computation of $\Sigma$ (Definition \ref{def:sigma}), the quantity closely related to the inverse Fisher Information.

\begin{lemma}
\label{lem:cov_passive}
\begin{align}
\Sigma_{\text{passive}} = \mathbb{E}[\sigma (1-\sigma) xx^\top]^{-1}
\end{align}
\end{lemma}
The proof is in the appendix. The proof relies on calibration, Assumption \ref{assum:calibrate}, to ensure that $\mathbb{E}[\nabla^2 \ell_w(x,y)] = \text{Cov}(\nabla \ell_w(x,y))$, which is always true for well-specified models.

This lemma gives $\Sigma$ as exactly the inverse Fisher information that was mentioned earlier. It is the expected value of $\nabla^2 \ell_{w^*}(x,y) = \sigma(1-\sigma)xx^\top$.

\begin{lemma}
\label{lem:cov_active}
\begin{align}
\Sigma_{\text{active}} = 
\end{align}
  $$\left((1-\alpha) \mathbb{E}_{x_1=0}[\sigma (1-\sigma) xx^\top] +  \alpha \mathbb{E}[\sigma (1-\sigma) xx^\top]\right)^{-1}$$
\end{lemma}

The proof is in the appendix. The proof relies on the assumptions of bounded support and Lipshitz pdf, Assumptions \ref{assum:bounded_support} and \ref{assum:lipschitz}. 

Because we randomly sample for $\alpha$ proportion, a factor of $\alpha$ times $\Sigma_{\text{passive}}$ shows up. Additionally, we get a $1-\alpha$ factor for the expected value of $\nabla^2 \ell_{w^*}(x,y) = \sigma(1-\sigma)xx^\top$ at the decision boundary. We will almost surely never sample exactly at the decision boundary, but as $n \rightarrow \infty$, the seed round weights $w_0 \rightarrow w^*$ and $\npool / n \rightarrow \infty$, we sample closer and closer to the decision boundary.

\subsubsection{Results}

Here, we define $s$ that quantifies how much the covariance at the decision boundary differs from the covariance for the rest of the distribution, which is a key dependency of our most general theorem.  Denote $x_{-1}$ as the vector $x$ without the first index. Recall that without loss of generality, $w^* = \|w^*\| e_1$.

\begin{definition}
  We define $s$ in terms of $C_0$ and $C_1$,
\begin{align}
C_0 &= \mathbb{E}_{x_1 = 0}[x_{-1} x_{-1}^\top] \\
C_1 &= \frac{\mathbb{E}[\sigma (1-\sigma) x_{-1} x_{-1}^\top]}{\mathbb{E}[\sigma (1-\sigma)]} \\
\frac{1}{s} &= \|C_0^{-1/2} C_1 C_0^{-1/2}\|_2
\end{align}

\end{definition}

We can give an interpretation to these constants. Define $C(a) = \mathbb{E}[x_{-1} x_{-1}^T | x_1=a]$ as the covariance of the directions orthogonal to $w^*$ at the slice $x_1=a$. Then, $C_0$ is simply $C(0)$, the covariance at the decision boundary. 

Further, define a variable $B$ that is $x_1$ weighted by $\sigma(1-\sigma)$:
\begin{align}
p(B=b) \propto \sigma(\|w^*\| b) (1-\sigma(\|w^*\| b)) p(x_1 = b).
\end{align}
Then, $C_1 = \mathbb{E}[C(B)]$, the covariance over the whole distribution, but weighted higher near the decision boundary with exponential tails. Finally, $s$ compares how much these two covariances differ.

Intuitively, we need this parameter to handle the case where the covariance at the decision boundary (a factor of the $\Sigma$ for active learning) is small relative to the average covariance.

Here is our main theorem which is proved by showing that
\begin{align}
\Sigma_{\text{passive},-1} \succ \frac{s}{4 \Err}\Sigma_{\text{active},-1}
\end{align}
and then using Lemma \ref{lem:rates}.

\begin{theorem}
\label{thm:main}
For sufficiently small constant $\alpha$ (that depends on the dataset) and for $Err < \epsilon < \epsilon_0$,
\begin{align}
DE(\epsilon) > \frac{s}{4 \Err}.
\end{align}
\end{theorem}

We can also get an upper bound on the data efficiency if we make an additional assumption that the pdf of $x$ factorizes into two
marginal distributions (decomposition of $x$ into two independent components),
one along the direction of $w^*$ and one in the directions orthogonal to $w^*$.

\begin{theorem}
\label{thm:decomposition_symmetric}
If $p(x) = p(x_1) p(x_{-1})$ and $p(x_1)=p(-x_1)$, then for sufficiently small constant $\alpha$ (that depends on the dataset), and for $\Err < \epsilon < \epsilon_0$,
\begin{align}
\frac{1}{4 \Err}  < DE(\epsilon) < \frac{1}{2 \Err}.
\end{align}
\end{theorem}

We can therefore see from these results that there is an inverse relationship between the asymptotic data efficiency and the population error, shedding light and giving a theoretical explanation to the empirical observation made in Section \ref{sec:experiments}.

\section{Discussion and Related Work}
\label{sec:discussion}

The conclusion of this work, that data efficiency is inversely related to limiting error, has been hinted at by a couple sentences in empirical survey papers. \citet{schein2007active} states ``the data sets sort neatly by noise, with [uncertainty] sampling failing on more noisy data ... and performing at least as well as random [sampling] for [less noisy] data sets.'' 
\citet{yang2016benchmark} states ``For the [less noisy] datasets, random sampling does not achieve the best performance ..., which may indicate that we need only consider random sampling on relatively [noisy] tasks''.

Additionally, this conclusion has evidence from statistical active learning theory \citep{hanneke2014statistical}. While not mentioned in the work, the ratio between the passive and active bounds points to a $1/\Err$ factor (though with $\Err$ being the optimal error over classifiers, not the MLE classifier). More specifically, the ratio between the passive and active lower bounds converges to $\Theta(1/\Err)$ as $\epsilon \rightarrow \Err$. Additionally, the ratio of active and passive algorithms converge to $\Theta(1/\Err)$; however with a factor of a disagreement coefficient which has a dimension dependence for linear classifiers and a $\log \log 1/\epsilon$ factor which ``is sometimes possible to remove'' \citep{hanneke2014statistical}.

This conclusion can be used in practice in at least two possible ways. First, a pilot study or domain knowledge can be used to get a rough estimate of the final error and if the error is low enough (less than around 10\%), uncertainty sampling can be used. Additionally, random sampling could be run until the test error is below 10\% and then a switch be made to uncertainty sampling.


Does our conclusion hold for other models? Because of the mathematical similarity to SVM, it's likely it also holds for hinge loss. It is possible that it also holds for neural networks with a a softmax layer, since the softmax layer is mathematically equivalent to logistic regression. In fact, \citet{geifman2017deep} performs experiments with deep neural networks and multiclass classification on MNIST (1\% error, 6x data efficiency), CIFAR-10 (10\%, 2x), and CIFAR-100 (35\%, 1x) and finds results that are explained well by our conclusion.

In conclusion, we make an observation, clearly define a phenomenon, demonstrate it empirically, and analyze it theoretically. The thesis of this work, that the data efficiency of uncertainty sampling on logistic regression is inversely proportional to the limiting error, sheds light on the appropriate use of active learning, enabling machine learning practitioners to intelligently choose their data collection techniques, whether active or passive.

\section*{Reproducibility}
The code, data, and experiments for this paper
are available on the CodaLab platform at

\href{https://worksheets.codalab.org/worksheets/0x8ef22fd3cd384029bf1d1cae5b268f2d/}{https://worksheets.codalab.org/worksheets/ 0x8ef22fd3cd384029bf1d1cae5b268f2d/}.

\section*{Acknowledgments}
\label{sec:acknowledgments}

This research was supported by NSF grant DGE-$1656518$.

\bibliography{bibliography}

\begin{thebibliography}{19}
\providecommand{\natexlab}[1]{#1}
\providecommand{\url}[1]{\texttt{#1}}
\expandafter\ifx\csname urlstyle\endcsname\relax
  \providecommand{\doi}[1]{doi: #1}\else
  \providecommand{\doi}{doi: \begingroup \urlstyle{rm}\Url}\fi

\bibitem[Balcan et~al.(2007)Balcan, Broder, and Zhang]{balcan2007margin}
Balcan, M.-F., Broder, A., and Zhang, T.
\newblock Margin based active learning.
\newblock In \emph{International Conference on Computational Learning Theory},
  pp.\  35--50. Springer, 2007.

\bibitem[Balcan et~al.(2009)Balcan, Beygelzimer, and
  Langford]{balcan2009agnostic}
Balcan, M.-F., Beygelzimer, A., and Langford, J.
\newblock Agnostic active learning.
\newblock \emph{Journal of Computer and System Sciences}, 75\penalty0
  (1):\penalty0 78--89, 2009.

\bibitem[Brinker(2003)]{brinker2003incorporating}
Brinker, K.
\newblock Incorporating diversity in active learning with support vector
  machines.
\newblock In \emph{Proceedings of the 20th International Conference on Machine
  Learning (ICML-03)}, pp.\  59--66, 2003.

\bibitem[Chaudhuri et~al.(2015)Chaudhuri, Kakade, Netrapalli, and
  Sanghavi]{chaudhuri2015convergence}
Chaudhuri, K., Kakade, S.~M., Netrapalli, P., and Sanghavi, S.
\newblock Convergence rates of active learning for maximum likelihood
  estimation.
\newblock In \emph{Advances in Neural Information Processing Systems}, pp.\
  1090--1098, 2015.

\bibitem[Freund et~al.(1997)Freund, Seung, Shamir, and
  Tishby]{freund1997selective}
Freund, Y., Seung, H.~S., Shamir, E., and Tishby, N.
\newblock Selective sampling using the query by committee algorithm.
\newblock \emph{Machine learning}, 28\penalty0 (2):\penalty0 133--168, 1997.

\bibitem[Geifman \& El-Yaniv(2017)Geifman and El-Yaniv]{geifman2017deep}
Geifman, Y. and El-Yaniv, R.
\newblock Deep active learning over the long tail.
\newblock \emph{arXiv preprint arXiv:1711.00941}, 2017.

\bibitem[Hanneke(2014)]{hanneke2014statistical}
Hanneke, S.
\newblock \emph{Statistical Theory of Active Learning}.
\newblock Now Publishers Incorporated, 2014.

\bibitem[Hoi et~al.(2009)Hoi, Jin, Zhu, and Lyu]{hoi2009semisupervised}
Hoi, S.~C., Jin, R., Zhu, J., and Lyu, M.~R.
\newblock Semisupervised svm batch mode active learning with applications to
  image retrieval.
\newblock \emph{ACM Transactions on Information Systems (TOIS)}, 27\penalty0
  (3):\penalty0 16, 2009.

\bibitem[Lewis \& Gale(1994)Lewis and Gale]{lewis1994sequential}
Lewis, D.~D. and Gale, W.~A.
\newblock A sequential algorithm for training text classifiers.
\newblock In \emph{Proceedings of the 17th annual international ACM SIGIR
  conference on Research and development in information retrieval}, pp.\
  3--12. Springer-Verlag New York, Inc., 1994.

\bibitem[Liu et~al.(2015)Liu, Reyzin, and Ziebart]{liu2015shift}
Liu, A., Reyzin, L., and Ziebart, B.~D.
\newblock Shift-pessimistic active learning using robust bias-aware prediction.
\newblock In \emph{AAAI}, pp.\  2764--2770, 2015.

\bibitem[Roy \& McCallum(2001)Roy and McCallum]{roy2001toward}
Roy, N. and McCallum, A.
\newblock Toward optimal active learning through monte carlo estimation of
  error reduction.
\newblock \emph{ICML, Williamstown}, pp.\  441--448, 2001.

\bibitem[Schein \& Ungar(2007)Schein and Ungar]{schein2007active}
Schein, A.~I. and Ungar, L.~H.
\newblock Active learning for logistic regression: an evaluation.
\newblock \emph{Machine Learning}, 68\penalty0 (3):\penalty0 235--265, 2007.

\bibitem[Schohn \& Cohn(2000)Schohn and Cohn]{schohn2000less}
Schohn, G. and Cohn, D.
\newblock Less is more: Active learning with support vector machines.
\newblock In \emph{ICML}, pp.\  839--846, 2000.

\bibitem[Settles(2010)]{settles2010active}
Settles, B.
\newblock Active learning literature survey.
\newblock \emph{Computer Sciences Technical Report}, 1648, 2010.

\bibitem[Seung et~al.(1992)Seung, Opper, and Sompolinsky]{seung1992query}
Seung, H.~S., Opper, M., and Sompolinsky, H.
\newblock Query by committee.
\newblock In \emph{Proceedings of the fifth annual workshop on Computational
  learning theory}, pp.\  287--294. ACM, 1992.

\bibitem[Sourati et~al.(2017)Sourati, Akcakaya, Leen, Erdogmus, and
  Dy]{sourati2017asymptotic}
Sourati, J., Akcakaya, M., Leen, T.~K., Erdogmus, D., and Dy, J.~G.
\newblock Asymptotic analysis of objectives based on fisher information in
  active learning.
\newblock \emph{Journal of Machine Learning Research}, 18\penalty0
  (34):\penalty0 1--41, 2017.

\bibitem[Tong \& Koller(2001)Tong and Koller]{tong2001support}
Tong, S. and Koller, D.
\newblock Support vector machine active learning with applications to text
  classification.
\newblock \emph{Journal of machine learning research}, 2\penalty0
  (Nov):\penalty0 45--66, 2001.

\bibitem[van~der Vaart(1998)]{vaart98asymptotic}
van~der Vaart, A.~W.
\newblock \emph{Asymptotic statistics}.
\newblock Cambridge University Press, 1998.

\bibitem[Yang \& Loog(2016)Yang and Loog]{yang2016benchmark}
Yang, Y. and Loog, M.
\newblock A benchmark and comparison of active learning for logistic
  regression.
\newblock \emph{arXiv preprint arXiv:1611.08618}, 2016.

\end{thebibliography}
\bibliographystyle{icml2018}

\clearpage
\onecolumn
\section{Appendix}

\subsection{Notation}

$w_0$ is the weights after the seed round.

$A_{-1}$ is the matrix without the first row and column.
$A_{1,-1}$ is the vector from the first row and all columns except the first column.

Generally, the $O(f(n))$ notation hides constants that only depend on the dataset, such as $\|w^*\|$, $s$, $B$, etc.

For the order of things going to zero, we first choose $\alpha$ to be small, then $r$ to be small, then $n$ to be large.

$w_0$ is weight vector after seed round

$$\epsilon_{\text{active}}(n) = \mathbb{E}_{f \sim \text{active},n \text{points}}[ Err(f) ]$$
$$\epsilon_{\text{passive}}(n) = \mathbb{E}_{f \sim \text{passive},n \text{points}}[ Err(f) ]$$

$$DE(\epsilon) = \frac{ \max \{n: \epsilon_{\text{passive}}(n) \geq \epsilon\}}{\max \{n: \epsilon_{\text{active}}(n) \geq \epsilon\}} = \frac{n_{passive}(\epsilon)}{n_{active}(\epsilon)}$$

Without loss of generality, assume $w^* = \|w^*\| e_1$, $w^*_0 = 0$, and $\mathbb{E}[x_{2:}]=0$.

With an abuse of notation, let $\sigma = \sigma(w^* \cdot x) = \sigma(\|w^*\| x_1)$.

\subsection{Losses}
Define $\sigma(x) = \frac{1}{1 + -\exp(x)}$.

The loss (negative log-likelihood) for a single data point under logistic regression is

$$l_w(x,y) = \log(1 + \exp(-w \cdot yx))$$

and so the gradient is

$$\nabla l_w(x,y) = - \frac{yx \exp(-w \cdot yx)}{1 + \exp(-w \cdot yx)} = - yx \sigma(-w \cdot yx)$$

and the Hessian is

$$\nabla^2 l_w(x,y) = \frac{(yx)(yx)^T \exp(w \cdot yx)}{(1 + \exp(w \cdot yx))^2}$$
$$ = \frac{xx^T}{(1 + \exp(w \cdot yx))(1 + \exp(-w \cdot yx))}$$
$$ = \sigma(w \cdot yx) \sigma(-w \cdot yx) xx^T$$

Note that $\sigma(-x) = 1-\sigma(x)$.


\subsection{Decision Boundary}

\begin{lemma}
\label{lemma:decision_boundary_prob}
For sufficiently small $r$, if $\|w' - w^*\|_2 \leq 2r$, then

$$|\int_{w'\cdot x = 0} p(x) - \int_{w^* \cdot x = 0} p(x)\| = O(r)$$
\end{lemma}
\begin{proof}
Without loss of generality (rotation and translation), let $w^*_0=0$, $w^*=\|w^*\|e_1$ and let $w' = w'_1 e_1 + w'_2 e_2$.

We sample from places where $w'_0 + w'_1 x_1 + w'_2 x_2 = 0$ which occurs when $x_1 = \frac{w'_2}{w'_1} x_2 + \frac{w'_0}{w'_1} = ax_2 + b$. From the theorem assumption, we know that $|w'_0|, |w'_2| \leq r$ and $|w'_1| \geq \|w^*\| - r \geq \frac{1}{2} \|w^*\|$ (for sufficiently small $r$) so we know that $|a|, |b| \leq O(r)$

Note that

$$|\int_{w'\cdot x = 0} p(x) - \int_{w^* \cdot x = 0} p(x)\| = \|\int_{x} p(x_1=ax + b,x_2=x) - p(x_1=0)|$$

(Note that the Jacobian of the change of variables has the following matrix which has determinant $1$)

$$\begin{bmatrix}
    1 & 0 \\
    -a & 1\\
  \end{bmatrix}
$$

$$|\int_{w'\cdot x = 0} p(x) - \int_{w^* \cdot x = 0} p(x)\| \leq \int_{x} |p(x_1=ax + b|x_2=x)p(x_2=x) - p(x_1=0|x_2=x)p(x_2=x)|$$

With the assumption that the conditional probabilities are Lipschitz,

$$ \leq \int_{x} L |ax + b| p(x_2=x)$$
$$ \leq a L B + b L$$
$$ = O(r)$$
\end{proof}

\begin{lemma}
\label{lemma:sample_close}
For sufficiently small $r$, if $\|w_0 - w^*\|_2 \leq r$, then with probability going to $1$ exponentially fast, all points from two-stage uncertainty sampling are from some hyperplane $w'$ such that $\|w' - w^*\| \leq 2r$.
\end{lemma}
\begin{proof}

For small enough $r$, then $\int_{w' \cdot x = 0} p(x) > p_0/2$ from the above lemma if $\|w_0 - w^*\|_2 \leq 2r$. Thus, the probability of an unlabeled point within the parallel plane with bias less than $r$ different from $w_0$ such that $\|w' - w_0\|_2 \leq r$ is at least $2 \frac{r}{\|w_0\|} (p_0/2) \geq \frac{r p_0}{2 \|w^*\|} = \Theta(r)$ (for sufficiently small $r$). 

Recall that $\npool = \omega(n)$ and $\nseed = o(n)$.

For sufficiently large $n$, the probability of at least $n$ points from the $\npool-\nseed$ unlabeled points falling in this range is

$$\Pr[ Binomial(\npool-\nseed, \text{probability of falling}) \geq n] \geq$$
$$\Pr[ Binomial(\npool/2,C_1 r)) \geq n ]$$

for some constant $C_1$. 

We can use a Chernoff bound (standard with $\delta=1/2$) since $\npool = \omega(n)$ to bound by $\exp(-\omega(n))$. Thus the probability that the planes we choose from are farther than $r$ away from $w_0$ goes to $0$ with rate faster than $\exp(-n)$.
\end{proof} 

\subsection{Convergence}

\begin{lem}[\ref{convergence}]
Both two-stage uncertainty sampling and random sampling converge to $w^*$.
\end{lem}
\begin{proof}
For passive learning, the Hessian of the population loss is positive definite because the data covariance is non-singular (Assumption \ref{assum:non_singular}). Thus, the population loss has a unique optimum. By the definition of $w^*$, $w^*$ is the minimizer. Since the sample loss converges to the population loss, the result of passive learning converges to $w^*$.

By a similar argument, the weight vector $w_0$ after the seed round converges to $w^*$ since $\nseed$ is super-constant (Assumption \ref{assum:seed}). Thus, for any $r>0$, with probability converging to $1$ as $n \rightarrow \infty$, $\|w_0 - w^*\| \leq r \leq \lambda/2$. By Lemma \ref{lemma:sample_close}, with probability going to $1$, all points selected are from hyperplanes $w$ where $\|w-w^*\| \leq 2r \leq \lambda$. Thus, by Assumption \ref{assum:convergence}, $\mathbb{E}_{w \cdot x = 0}[\nabla l_{w^*}(x,y)] = 0$. In the second stage, because of the $\alpha$ proportion of randomly selected points, the loss from the new uncertainty sampling population has a unique optimum. And because the expectation of the gradient of the loss is $0$ for the points near the decision boundary (with probability going to $1$), the result of two-stage uncertainty sampling converges in probability to $w^*$.
\end{proof}

\subsection{Rates}

\begin{lem}
\label{lem:ck}
If $\Sigma$ exists, and for any $\epsilon>0$, $n \Pr[\|A_n - A\| \geq \epsilon] \rightarrow 0$ and $n \Pr[\|w_n - w^*\| \geq \epsilon] \rightarrow 0$, then there exist vectors $c_k \neq 0$ that depend only on the data distribution such that,

$$n(\epsilon(n) - Err) \rightarrow \sum_k c_k^T \Sigma_{-1} c_k$$
\end{lem}
\begin{proof}
The zero-one error is 

$$Z(w_n) = \Pr[yx \cdot w_n < 0]$$

Since $Z$ is twice differentiable at $w^*$, by Taylor's theorem,

$$Z(w_n) = Z(w^*) + (\nabla Z(w^*))^T (w_n-w^*) + (w_n-w^*)^T (\frac{1}{2} \nabla^2 Z(w^*)) (w_n-w^*) + (w_n-w^*)^T R(w_n - w^*) (w_n-w^*)^T$$

where $R(w) \rightarrow 0$ as $w \rightarrow 0$.

Since $Z$ has a local optimum at $w^*$, $\nabla Z(w^*)=0$. Also $Z(w^*) = Err$. Additionally, denote $H = \frac{1}{2} \nabla^2 Z(w^*)$,

$$Z(w_n) = Err + (w_n-w^*)^T (H + R(w_n - w^*)) (w_n-w^*)$$

Choose any $\epsilon>0$.  Since $R(w) \rightarrow 0$ as $w \rightarrow 0$, there is $\delta_\epsilon$ such that $\|w\| \leq \delta_\epsilon \implies \|R(w)\| \leq \epsilon$. Define $near(n)$ to be the event that $\|A_n - A\| \geq \epsilon \wedge \|w_n - w^*\| \geq \delta_\epsilon$. Note that from the theorem assumption, $n \Pr[\neg near(n)] \rightarrow 0$. 

$$\epsilon(n) = \mathbb{E}[Z(w_n)] = \Pr[\neg near(n)] \mathbb{E}[Z(w_n)|\neg near(n)] + \Pr[near(n)] \mathbb{E}[Z(w_n)|near(n)]$$

$$|n \epsilon(n) - n \mathbb{E}[Z(w_n)|near(n)]| \leq n \Pr[\neg near(n)] |\mathbb{E}[Z(w_n)|\neg near(n)] - \mathbb{E}[Z(w_n)|near(n)]|$$
$$\leq n \Pr[\neg near(n)] \rightarrow 0$$

Thus, 

$$n(\epsilon(n) - Err) \rightarrow n(\mathbb{E}[Z(w_n)|near(n)] - Err)$$

So we need to just worry about the convergence of the right side,

$$\mathbb{E}[Z(w_n)|near(n)] = Err + \frac{1}{n} \mathbb{E}[ (A_n^{-1} b_n)^T (H + R(w_n - w^*)) (A_n^{-1} b_n) | near(n)]$$

$$n(\mathbb{E}[Z(w_n)|near(n)]- Err) = \mathbb{E}[ b_n^T A_n^{-1} (H+R(w_n-w^*)) A_n^{-1} b_n | near(n)]$$

Because we conditioned on $near(n)$, $\|A_n - A\| \leq \epsilon$ and $\|w_n - w^*\| \leq \delta_\epsilon$ and therefore $\|R(w_n - w^*)\| \leq \epsilon$. So $\|A_n^{-1} (H+R(w_n-w^*)) A_n^{-1} - A^{-1} H A^{-1}\| = O(\epsilon)$. Using this, we get,

$$\|n(\mathbb{E}[Z(w_n)|near(n)]- Err) - \mathbb{E}[ b_n^T A^{-1} H A^{-1} b_n | near(n)]\| \leq \|\mathbb{E}[ b_n^T O(\epsilon) b_n | near(n)]\|$$
$$ \leq O(\epsilon) \|\mathbb{E}[ \|b_n\|^2 | near(n)]\|$$
$$ \leq O(\epsilon) \|\mathbb{E}[ b_n b_n^T | near(n)]\|$$

Note that,
$$\mathbb{E}[b_n b_n^T] = \mathbb{E}[b_n b_n^T | near(n)] \Pr[near(n)] + \mathbb{E}[b_n b_n^T | \neg near(n)] \Pr[\neg near(n)]$$
and the later two expectations exist since the left exists and the matrices are positive semidefinite. Passing through the limit, we see that $\mathbb{E}[b_n b_n^T | near(n)] \rightarrow B$. 

Thus, noting that we can drive $\epsilon \rightarrow 0$,

$$n(\mathbb{E}[Z(w_n)|near(n)]- Err) \rightarrow \mathbb{E}[ b_n^T A^{-1} H A^{-1} b_n | near(n)]$$
$$\rightarrow \sum_{i,j} [A^{-1} H A^{-1}]_{i,j} \mathbb{E}[b_n b_n^T| near(n)]_{i,j}$$
$$\rightarrow \sum_{i,j} [A^{-1} H A^{-1}]_{i,j} B_{i,j}$$

Thus, putting this together, we see that

$$n(\epsilon(n) - Err) \rightarrow \sum_{i,j} [A^{-1} H A^{-1}]_{i,j} B_{i,j}$$

Doing manipulations on the indices, we find,

$$\sum_{i,j} [A^{-1} H A^{-1}]_{i,j} B_{i,j} = \sum_{i,j} H_{i,j} (A^{-1} B A^{-1})_{i,j}$$
$$= \sum_{i,j} H_{i,j} \Sigma_{i,j}$$

Therefore, 

$$n(\epsilon(n) - Err) \rightarrow \sum_{i,j} H_{i,j} \Sigma_{i,j}$$

and we are most of the way there, just need to use some properties to show the final form.

Since $w^*$ is a local optimum, $H \succeq 0$ (and symmetric) and since the Hessian is not identically zero at $w^*$, $H \neq 0$.

Without loss of generality, let $w^* = \|w^*\|e_1$ and $w^*_0 = 0$ as assumed before. Note that $Z(w^* + \alpha e_1) = Z(w^*)$ for $\alpha \in (-\|w^*\|/2, \infty)$. Since it is constant along this line, $(\nabla^2 Z(w^*))_{1,1}=0$, and so $H_{1,1}=0$

So $H \succeq 0$, $H$ is symmetric, $H \neq 0$, and $H_{1,1}=0$. Since $H \succeq 0$ and $H_{1,1}=0$, $H_{1,i}=0$ for all $i$.

Since $H \succeq 0$ and $H \neq 0$,

$H = \sum_k c_k c_k^T$

for some vectors $c_k$ (where there is at least one). And further, $(c_k)_1=0$.

$$\sum_{i,j} H_{i,j} \Sigma_{i,j} = \sum_{i,j} (\sum_k c_k c_k^T)_{i,j} \Sigma_{i,j}$$
$$ = \sum_k c_k^T \Sigma c_k$$

We can remove the first elements of $c_k$ and the first row and column of $\Sigma$ without changing anything, so 

$$\sum_{i,j} H_{i,j} \Sigma_{i,j} = \sum_k c_k^T \Sigma_{-1} c_k$$

And thus the theorem is proved.

\end{proof}

\begin{lem}
\label{lem:rates_corollary}
If we have two algorithms $a$ and $b$ that satisfy the conditions of Lemma \ref{lem:ck}, and 

$$\Sigma_{a,-1} \succ c \Sigma_{b,-1}$$

then there exists $\epsilon_0$ such that for $Err<\epsilon<\epsilon_0$,

$$n_a(\epsilon) \geq c n_b(\epsilon) $$
\end{lem}
\begin{proof}

$$\Sigma_{a,-1} \succ \alpha \Sigma_{b,-1}$$

$$\sum_k c_k^T \Sigma_{a,-1} c_k > \alpha \sum_k c_k^T \Sigma_{b,-1} c_k$$

so, for $n>n_0, n'>n_0$,

$$n(\epsilon_a(n) - Err) > \alpha n'(\epsilon_b(n') - Err)$$

setting $n' = n/\alpha$ and for $n>\max(n_0,n_0/\alpha)$,

$$n(\epsilon_a(n) - Err) > n(\epsilon_b(n / \alpha) - Err)$$

So for sufficiently large $n$,

$$\epsilon_a(n) > \epsilon_b(n / \alpha)$$

For any $\epsilon>Err$ such that $n_a(\epsilon)$ is sufficiently large, (we know this exists since $n_a(\epsilon) = \Theta(\frac{1}{\epsilon - Err})$)

$$\epsilon_a(n) \leq \epsilon     \text{  for  }     n \geq n_a(\epsilon)$$
$$\epsilon_b(n/\alpha) \leq \epsilon \text{  for  } n \geq n_a(\epsilon)$$
$$\epsilon_b(n') \leq \epsilon \text{  for  }   n' \geq \frac{1}{\alpha} n_a(\epsilon)$$
$$n_b(\epsilon) \leq \frac{1}{\alpha} n_a(\epsilon)$$
$$n_a(\epsilon) \geq \alpha n_b(\epsilon) $$

\end{proof}

\begin{lem}[\ref{lem:rates}]
If we have two algorithms with $\Sigma_a$ and $\Sigma_b$, and for any $\epsilon>0$ and both estimators, $n \Pr[\|A_n - A\| \geq \epsilon] \rightarrow 0$ and $n \Pr[\|w_n - w^*\| \geq \epsilon] \rightarrow 0$, then 

$$\Sigma_{a,-1} \succ c \Sigma_{b,-1}$$

implies that for some $\epsilon_0$ and any $Err<\epsilon<\epsilon_0$,

$$n_a(\epsilon) \geq c n_b(\epsilon) $$
\end{lem}
\begin{proof}
This is a straightforward application of the above lemmas, Lemma \ref{lem:ck} and Lemma \ref{lem:rates_corollary}.
\end{proof}

\subsection{Conditions satisfied}

\begin{lem}[\ref{lem:ck_conditions}]
For our active and passive learning algorithms,
for any $\epsilon>0$, $n \Pr[\|A_n - A\| \geq \epsilon] \rightarrow 0$ and $n \Pr[\|w_n - w^*\| \geq \epsilon] \rightarrow 0$
\end{lem}
\begin{proof}
Recall that 

$$A_n = \frac{1}{n} \sum_i \nabla^2 l_{w'}(x_i,y_i)$$  
$$b_n = \frac{1}{\sqrt{n}} \sum_i \nabla l_{w^*}(x_i,y_i)$$

where $\|w' - w^*\| \leq \|w_n - w^*\|$.

For passive learning, by CLT, for any $\epsilon$, $\Pr[\|w_n - w^*\| > \epsilon] = O(\frac{e^{-\Theta(n)}}{\sqrt{n}})$. Thus, we find that $n \Pr[\|w_n - w^*\| \geq \epsilon] \rightarrow 0$.

We also need this fact to bound $w'$. Then, with a Hoeffding bound on the sum of $A_n$, we can get that $\Pr[\|A_n - A\| \geq \epsilon] = O(\frac{e^{-\Theta(n)}}{\sqrt{n}})$ and thus  $n \Pr[\|A_n - A\| \geq \epsilon] \rightarrow 0$.

For active learning, we need to be careful because if $\|w_0 - w^*\| > \lambda/2$, we are not even guaranteed that the final result converges (see Lemma \ref{lemma:sample_close}). However, by the CLT, we find that $\Pr[\|w_0 - w^*\| > \lambda/2] = O(\frac{e^{-\Theta(\nseed)}}{\sqrt{\nseed}})$. Because $\nseed=\Omega(n^\rho)$ (see Assumption \ref{assum:seed}), this converges exponentially fast and $n \Pr[\|w_0 - w^*\| > \lambda/2] \rightarrow 0$.

Because of the $\alpha$ random sampling, and conditioned on the probability that $\|w_0 - w^*\| < \lambda/2$, we can get the same results for active learning as for passive learning. Note that from Lemma \ref{lemma:sample_close}, there is exponentially small probability of not sampling all points from $w'$ where $\|w' - w^*\| < \lambda$.

\end{proof}

\subsection{$COV$ calculation for passive}

\begin{lemma}
\label{lem:calibrate_B}
For passive learning, $\mathbb{E}[\nabla l_{w^*}(x,y) (\nabla l_{w^*}(x,y))^T] = \mathbb{E}[\sigma (1-\sigma) xx^T]$. 
\end{lemma}
\begin{proof}
Since the mean of the derivative of the loss is $0$ at $w^*$,

$$\mathbb{E}[\nabla l_{w^*}(x,y) (\nabla l_{w^*}(x,y))^T]_{i,j} =  \mathbb{E}[x_i x_j \sigma(-\|w^*\| yx_1)^2]$$

$$ = \mathbb{E}_{x_1}[  \mathbb{E}[x_i x_j| x_1] \mathbb{E}[\sigma(\|w^*\| yx_1)^2|x_1]]$$
$$ = \mathbb{E}_{x_1}[  \mathbb{E}[x_i x_j| x_1] [ P(y=1 | x_1) \sigma(-\|w^*\| x_1)^2 + P(y=1 | x_1) \sigma(\|w^*\| x_1)^2]]$$

from the calibrated assumption,

$$ = \mathbb{E}_{x_1}[  \mathbb{E}[x_i x_j| x_1] [ \sigma(\|w^*\| x_1) \sigma(-\|w^*\| x_1)^2 + \sigma(-\|w^*\| x_1) \sigma(\|w^*\| x_1)^2]]$$
$$ = \mathbb{E}_{x_1}[  \mathbb{E}[x_i x_j| x_1] \sigma(\|w^*\| x_1) \sigma(-\|w^*\| x_1) [ \sigma(\-|w^*\| x_1) + \sigma(\|w^*\| x_1)]]$$
$$ = \mathbb{E}_{x_1}[  \mathbb{E}[x_i x_j| x_1] \sigma(\|w^*\| x_1) \sigma(-\|w^*\| x_1)]$$
$$ = \mathbb{E}[x_i x_j \sigma(\|w^*\| x_1) \sigma(-\|w^*\| x_1)]$$
$$ = \mathbb{E}[\sigma(1-\sigma)xx^T]_{i,j}$$
\end{proof}

\begin{lem}[\ref{lem:cov_passive}]
$$\Sigma_{passive} = [\mathbb{E}[\sigma (1-\sigma) xx^T]]^{-1}$$
\end{lem}
\begin{proof}
For passive learning, by the convergence of $w^n \rightarrow w^*$ and by the law of large numbers,

$$A_n \rightarrow A = \mathbb{E}[\sigma (1-\sigma) xx^T ]$$

Further, by independence of draws,

$$\mathbb{E}[b_n b_n^T] = \mathbb{E}[\nabla l_{w^*}(x,y) (\nabla l_{w^*}(x,y))^T]$$
so by Lemma \ref{lem:calibrate_B},
$$\mathbb{E}[b_n b_n^T] = \mathbb{E}[\sigma (1-\sigma) xx^T]$$
$$B = \mathbb{E}[\sigma (1-\sigma) xx^T]$$
$$B = A$$

Thus,

$$\Sigma_{passive} = A^{-1} B A^{-1}$$
$$ = A^{-1}$$
$$ = [\mathbb{E}[\sigma (1-\sigma) xx^T]]^{-1}$$

\end{proof}

\subsection{$COV$ calculation for active}

\begin{lemma}
\label{lem:active_plane_converge}
For sufficiently small $r$ (small with respect to dataset-only dependent constants), if $\|w' - w^*\|_2 \leq 2 r$, then

$$\| \mathbb{E}_{w' \cdot x = 0}[\sigma (1-\sigma) xx^T] - \mathbb{E}_{w^* \cdot x = 0}[\sigma (1-\sigma) xx^T] \| = O(r)$$

and

$$\| \mathbb{E}_{w' \cdot x = 0}[\sigma(-y x_1 \|w^*\|)^2 xx^T] - \mathbb{E}_{w^* \cdot x = 0}[\sigma(-y x_1 \|w^*\|)^2 xx^T] \| = O(r)$$

\end{lemma}
\begin{proof}
Without loss of generality (rotation and translation), let $w^*_0=0$, $w^*=\|w^*\|e_1$ and let $\hat{w} = c_1 e_1 + c_2 e_2$.

We sample from places where $w'_0 + w'_1 x_1 + w'_2 x_2 = 0$ which occurs when $x_1 = \frac{w'_2}{w'_1} x_2 + \frac{w'_0}{w'_1} = ax_2 + b$. From the theorem assumption, we know that $|w'_0|, |w'_2| \leq r$ and $|w'_1| \geq \|w^*\| - r \geq \frac{1}{2} \|w^*\|$ (for sufficiently small $r$) so we know that $|a|, |b| \leq O(r)$

Define $Q(x_1) = \sigma(\|w^*\| x_1) \sigma(-\|w^*\| x_1)$ or $Q(x_1) = \sigma(- y x_1 \|w^*\|)^2$ (abuse of notation). Both these functions are Lipschitz around $x_1=0$, and bounded (since support bounded by $B$).

First, we compute the joint (not the conditionals) and then we can divide by the marginals from the previous lemma,

Let $i_1, i_2, ..., i_d$ be indicators for the indices $i,j$ that are non-zero. Thus, $i_1 + i_2 + ... + i_d \leq 2$,

$$\mathbb{E}_{w' \cdot x = 0}[\sigma (1-\sigma) xx^T]_{i,j} = $$

$$=\mathbb{E}_{w' \cdot x = 0}[Q(x_1) (x_1)^{i_1} (x_2)^{i_2} (x_3)^{i_3} ... ] = $$

(As before, the Jacobian of the change of variables has determinant $1$)
$$\int_{x} p(x_1 = ax + b, x_2=x) Q(ax + b) (ax + b)^{i_1} (x)^{i_2} \mathbb{E}[x_3^{i_3}  ... | x_1 = ax + b, x_2=x]=$$ 
$$ = \int_{x} p(x_2=x) (x)^{i_2} F(ax+b, x)$$

where $F(x_1, x_2) = p(x_1 | x_2)  (Q(x_1) x_1^{i_1}) mathbb{E}[x_3^{i_3}  ... | x_1, x_2]$

All three components of $F$ are bounded, since support bounded, Assumption \ref{assum:bounded_support}. Further, all three components are Lipschitz, because of Assumption \ref{assum:lipschitz} and bounded support as well. Therefore, $F$ is Lipschitz.

$$|\int_{x} p(x_2=x) (x)^{i_2} F(ax+b,x) - \int_{x} p(x_2=x) (x)^{i_2} F(0,x) $$
$$\leq  \int_{x} p(x_2=x) |x|^{i_2} L |ax+b|$$
$$ \leq  a L B^{i_2+1} + b L B^{i_2}$$
$$ = O(r)$$

Thus, for any $i,j$, 

$$\| \mathbb{E}_{w' \cdot x = 0}[Q xx^T]_{i,j} - \mathbb{E}_{w^* \cdot x = 0}[Q xx^T]_{i,j} \| = O(r)$$

We can use this to bound the matrix norm,

$$\| \mathbb{E}_{w' \cdot x = 0}[Q xx^T] - \mathbb{E}_{w^* \cdot x = 0}[Q xx^T] \| = O(r)$$

Since the probabilities (see Lemma \ref{lemma:decision_boundary_prob})  and conditionals are both off by only $O(r)$ (from above) and since the probabilities are bounded away from $0$ (see Lemma \ref{lemma:decision_boundary_prob} and Assumption \ref{assum:non_zero_prob}), the conditional distribution is off by $O(r)$. We can plug in both functions of $Q$ to get the statement of the theorem.
\end{proof}


\begin{lem}[\ref{lem:cov_active}]
$$\Sigma_{active} = [(1-\alpha) \mathbb{E}_{x_1=0}[\sigma (1-\sigma) xx^T] +  \alpha \mathbb{E}[\sigma (1-\sigma) xx^T]]^{-1}$$
\end{lem}
\begin{proof}

Because $w_n \rightarrow w^*$, and by the law of large numbers,

$$A_n \rightarrow (1-\alpha) \mathbb{E}_{w'}[\mathbb{E}_{w' \cdot x = 0}[\sigma(-y x_1 \|w^*\|)^2 xx^T]] + \alpha  \mathbb{E}[\sigma(-y x_1 \|w^*\|)^2 xx^T]$$

From Lemma \ref{lem:active_plane_converge},

$$\| \mathbb{E}_{w' \cdot x = 0}[\sigma (1-\sigma) xx^T] - \mathbb{E}_{w^* \cdot x = 0}[\sigma (1-\sigma) xx^T] \| = O(r)$$

and $\|w'-w^*\| < 2r$ with probability going to $1$,

$$A_n \rightarrow \frac{n - \nseed}{n} [(1-\alpha) \mathbb{E}_{w^* \cdot x = 0}[\sigma (1-\sigma) xx^T] + O(r) + \alpha  \mathbb{E}[\sigma (1-\sigma) xx^T]]$$

Since $w_0 \rightarrow w^*$, $r \rightarrow 0$, and since $\nseed = o(n)$ (see Assumption \ref{assum:seed}) so 

$$A_n \rightarrow A = (1-\alpha) \mathbb{E}_{w^* \cdot x = 0}[\sigma (1-\sigma) xx^T] + \alpha  \mathbb{E}[\sigma (1-\sigma) xx^T]$$

The same line of argument with using Lemma \ref{lem:active_plane_converge} and Lemma \ref{lem:calibrate_B} yields

$$B = A$$

So

$$\Sigma_{active} = A^{-1} B A^{-1} = A^{-1}$$
$$ = [(1-\alpha) \mathbb{E}_{x_1 = 0}[\sigma (1-\sigma) xx^T] + \alpha  \mathbb{E}[\sigma (1-\sigma) xx^T]]^{-1}$$

\end{proof}

\subsection{Inverses Without First Coordinate}

\begin{lemma}

$$\begin{bmatrix}
    a & \vec{a}^T \\
    \vec{a} & A \\
  \end{bmatrix}^{-1} =  
  \begin{bmatrix}
    b & \vec{b}^T \\
    \vec{b} & B \\
  \end{bmatrix}
$$

Where

$$b = \frac{1}{a - \vec{a}^T A^{-1} \vec{a}}$$
$$\vec{b} = -b A^{-1} \vec{a}$$
$$B = A^{-1} + b (A^{-1} \vec{a}) (A^{-1} \vec{a})^T$$
\end{lemma}
\begin{proof}
Matrix algebra.
\end{proof}

\begin{lemma}
\label{lem:inverse_without_first}
$$(A^{-1})_{-1} = (A_{-1})^{-1} + \frac{((A_{-1})^{-1} A_{-1,1}) ((A_{-1})^{-1} A_{-1,1})^T }{A_{1,1} - A_{-1,1}^T (A_{-1})^{-1} A_{-1,1}}$$
\end{lemma}
\begin{proof}
Use the above theorem and note that $b>0$ so 

$$b (A^{-1} \vec{a}) (A^{-1} \vec{a})^T \succeq 0$$
\end{proof}
\subsection{Relating Err to expectation of sigmoid}

\begin{lemma}
\label{lem:err_relate}
$$\frac{Err}{2} < \mathbb{E}[\sigma(1-\sigma)] < Err$$
\end{lemma}
\begin{proof}

$$Err = P(y x_1 \|w^*\| < 0)$$
$$ = P(x_1 < 0 \wedge y=1) + P(x_1 > 0 \wedge y=-1)$$

From Assumption \ref{assum:calibrate},

$$ = \int_{-\infty}^{0} p_{x_1}(x_1) \sigma(-w^*_1 x_1) + \int_{0}^{0\infty} p_{x_1}(x_1) \sigma(w^*_1 x_1)$$
$$ = \int_{0}^{\infty} [p_{x_1}(-x_1) + p_{x_1}(x_1)] \sigma(w^*_1 x_1) $$

Additionally,

$$\mathbb{E}[\sigma(1-\sigma)] = \mathbb{E}[\sigma(y x_1 \|w^*\|) \sigma(-y x_1 \|w^*\|)]$$
$$ = \mathbb{E}[\sigma(\|w^*\| x_1) \sigma(-\|w^*\| x_1)]$$
$$ = \int_{-\infty}^{0} p_{x_1}(x_1) \sigma(\|w^*\| x_1) \sigma(-\|w^*\| x_1) + \int_{0}^{\infty} p_{x_1}(x_1) \sigma(\|w^*\| x_1) \sigma(-\|w^*\| x_1)$$
$$ = \int_{0}^{\infty} [p_{x_1}(-x_1) + p_{x_1}(x_1)] \sigma(\|w^*\| x_1) \sigma(-\|w^*\| x_1)$$

Note that for $x_1 > 0$, $\frac{1}{2} < \sigma(-\|w^*\| x_1) < 1$. Comparing equations, we get,

$$\frac{Err}{2} < \mathbb{E}[\sigma(1-\sigma)] < Err$$
\end{proof}

\subsection{Main DE bound}

\begin{thmn}[\ref{thm:main}]
For sufficiently small constant $\alpha$ (that depends on the dataset) and for $Err < \epsilon < \epsilon_0$,

$$DE(\epsilon) > \frac{s}{4 Err}$$
\end{thmn}
\begin{proof}
For convenience, define 

$$Q = \mathbb{E}_{x_1=0}[\sigma (1-\sigma) xx^T]$$
$$R = \mathbb{E}[\sigma (1-\sigma) xx^T] = COV_{passive}$$
$$S = \alpha R + (1-\alpha) Q = COV_{active}$$

By the definition of $s$,

$$\mathbb{E}_{x_1=0}[x_{-1}x_{-1}^T] \succeq s \frac{\mathbb{E}[\sigma (1-\sigma) x_{-1}x_{-1}^T]}{\mathbb{E}[\sigma (1-\sigma)]}$$

By Lemma \ref{lem:err_relate},

$$4 Q_{-1} \succ \frac{s}{Err} R_{-1}$$

For small enough $\alpha$,

$$Q_{-1} \succ \frac{s/(4 Err) - \alpha}{1 - \alpha} R_{-1}$$

$$\alpha R_{-1} + (1-\alpha)Q_{-1} \succ \frac{s}{4 Err} R_{-1}$$

$$S_{-1} \succ \frac{s}{4 Err} R_{-1}$$

$$ \frac{s}{4 Err} (S_{-1})^{-1} \prec (R_{-1})^{-1} \preceq (R^{-1})_{-1}$$

The last step comes from noting that the right hand side of Lemma \ref{lem:inverse_without_first} positive semidefinite for $A$ positive semidefinite.

Additionally, note that the first row and column of $Q$ is $0$,

so $S_{-1,1} = \alpha R_{-1,1}$ and $S_{1,1} = \alpha R_{1,1}$.

An examination yields,

$$  \frac{(S_{-1})^{-1} S_{-1,1}) (S_{-1})^{-1} S_{-1,1})^T }{S_{1,1} - S_{-1,1}^T (S_{-1})^{-1} S_{-1,1}} = O(\alpha)$$

Using Lemma \ref{lem:inverse_without_first}, we find that we can make $\alpha$ small enough so that

$$\frac{s}{4 Err} (S^{-1})_{-1} \prec  (R^{-1})_{-1}$$

$$\frac{s}{4 Err} COV_{active,-1} \prec  COV_{passive,-1}$$

so by Lemma \ref{lem:rates}, for $Err< \epsilon < \epsilon_0$,

$$DE(\epsilon) > \frac{s}{4 Err}$$

\end{proof}
\subsection{DE Bound Given Decomposition}

We actually get a slightly more general result from the following lemma.

\begin{lemma}
\label{lem:decomposition}
If $p(x) = p(x_1) p(x_{-1})$, then for sufficiently small constant $\alpha$ (that depends on the dataset), and for $Err < \epsilon < \epsilon_0$,

$$\frac{1}{4 Err}  < DE(\epsilon) < \frac{1}{2 Err} (1 + \frac{\mathbb{E}[\widetilde{X}]}{Var(\widetilde{X})})$$

where 

$$p(\widetilde{X}=x) \propto \sigma(\|w^*\| x) (1-\sigma(\|w^*\| x)) p(x_1 = x)$$
\end{lemma}
\begin{proof}
With the decomposition, in the Theorem \ref{thm:main}, $s=1$. So we get for free that for $Err < \epsilon < \epsilon_0$,

$$DE(\epsilon) > \frac{1}{4 Err}$$

As before, for convenience, define

$$Q = \mathbb{E}_{x_1=0}[\sigma (1-\sigma) xx^T]$$
$$R = \mathbb{E}[\sigma (1-\sigma) xx^T] = COV_{passive}$$
$$S = \alpha R + (1-\alpha) Q = COV_{active}$$

Because of the decomposition,

$$R_{2:,2:} = \mathbb{E}[\sigma(1-\sigma)] \mathbb{E}[x_{2:} x_{2:}^T] \succ \frac{Err}{2} \mathbb{E}[x_{2:} x_{2:}^T]$$
$$Q_{2:,2:} = \frac{1}{4} \mathbb{E}[x_{2:} x_{2:}^T]$$
$$Q_{2:,2:} \prec \frac{1}{2 Err} R_{2:,2:}$$
For sufficiently small $\alpha$,
$$Q_{2:,2:} \prec \frac{1/(2 Err) - \alpha}{1 - \alpha} R_{2:,2:}$$
$$\alpha R_{2:,2:} + (1 - \alpha) Q_{2:,2:} \prec \frac{1}{2 Err} R_{2:,2:}$$
$$S_{2:,2:} \prec \frac{1}{2 Err} R_{2:,2:}$$

Because of the decomposition, and because $\mathbb{E}[x_{2:}]=0$ (without loss of generality by translation),

$$R_{0:1,2:} = 0$$
$$Q_{0:1,2:} = 0$$

$$\frac{1}{2 Err} (A^{-1})_{2:,2:} \succ  (R^{-1})_{2:,2:}$$

Now, let us examine the upper left corners,

$$R_{0:1,0:1} = 
\begin{bmatrix}
  \mathbb{E}[\sigma(1-\sigma)] & \mathbb{E}[\sigma(1-\sigma) x_1] \\
  \mathbb{E}[\sigma(1-\sigma) x_1] & \mathbb{E}[\sigma(1-\sigma) x_1^2] \\
\end{bmatrix}
$$
$$S_{0:1,0:1} =
\begin{bmatrix}
  (1-\alpha)/4 + \alpha \mathbb{E}[\sigma(1-\sigma)] & \alpha \mathbb{E}[\sigma(1-\sigma) x_1] \\
  \alpha \mathbb{E}[\sigma(1-\sigma) x_1] & \alpha \mathbb{E}[\sigma(1-\sigma) x_1^2] \\
\end{bmatrix}
$$

Denote

$$D = \mathbb{E}[\sigma(1-\sigma)] \mathbb{E}[\sigma(1-\sigma) x_1^2] - \mathbb{E}[\sigma(1-\sigma) x_1]^2$$

Then,

$$(R^{-1})_{0,0} = \frac{\mathbb{E}[\sigma(1-\sigma) x_1^2]}{D}$$
$$(S^{-1})_{0,0} = \frac{\alpha \mathbb{E}[\sigma(1-\sigma) x_1^2]}{\alpha (1-\alpha) (1/4) \mathbb{E}[\sigma(1-\sigma) x_1^2] + \alpha^2 D}$$

$$(R^{-1})_{0,0} / (S^{-1})_{0,0} = \frac{1-\alpha}{4 \mathbb{E}[\sigma(1-\sigma)]} (1+\frac{\mathbb{E}[\sigma(1-\sigma) x_1]^2}{D}) + \alpha$$

For small enough $\alpha$,

$$(R^{-1})_{0,0} / (S^{-1})_{0,0} < \frac{1}{2 Err} (1+\frac{\mathbb{E}[\sigma(1-\sigma) x_1]^2}{D})$$

Combining the bounds on the two blocks of the matrices, we get that

$$\frac{1}{2 Err} (1+\frac{\mathbb{E}[\sigma(1-\sigma) x_1]^2}{D}) (S^{-1})_{-1} \succ (R^{-1})_{-1}$$

$$\frac{1}{2 Err} (1+\frac{\mathbb{E}[\sigma(1-\sigma) x_1]^2}{D}) COV_{active,-1} \succ COV_{passive,-1}$$

So for $\epsilon < \epsilon_0$,

$$DE(\epsilon) < \frac{1}{2 Err} (1+\frac{\mathbb{E}[\sigma(1-\sigma) x_1]^2}{D})$$

if we define $\widetilde{X}$ such that 
$p_{\widetilde{X}}(x) \propto \sigma (1-\sigma) p_{x_1}(x)$,

$$DE(\epsilon) < \frac{1}{2 Err} (1+\frac{\mathbb{E}[\widetilde{X}]^2}{Var(\widetilde{X})})$$
\end{proof}

\begin{thmn}[\ref{thm:decomposition_symmetric}]
If $p(x) = p(x_1) p(x_{-1})$ and $p(x_1)=p(-x_1)$, then for sufficiently small constant $\alpha$ (that depends on the dataset), and for $\Err < \epsilon < \epsilon_0$,

$$\frac{1}{4 \Err}  < DE(\epsilon) < \frac{1}{2 \Err}$$
\end{thmn}
\begin{proof}
If $p(x_1)=p(-x_1)$, then $p(\widetilde{X}) = p(-\widetilde{X})$ and so $\mathbb{E}[\widetilde{X}]=0$.

Using Lemma \ref{lem:decomposition}, we arrive at the conclusion.
\end{proof}

\end{document}